\documentclass{llncs}

\usepackage[english]{babel}
\usepackage[utf8]{inputenc}
\usepackage{latexsym,amssymb,amstext,amsfonts,amsmath,graphicx,setspace,mathtools,bm}
\usepackage{booktabs}
\usepackage[colorinlistoftodos]{todonotes}
\usepackage{enumerate} 
\usepackage{paralist}
\usepackage{float}
\usepackage{array}
\usepackage{comment}
\usepackage[draft]{hyperref}
\usepackage[font=scriptsize,labelfont=scriptsize,bf,skip=1pt]{caption}
\usepackage[font=scriptsize,labelfont=scriptsize]{subcaption}

\makeatletter
\renewcommand*{\@fnsymbol}[1]{\ensuremath{\ifcase#1\or \dagger\or *\or \ddagger\or
    \mathsection\or \mathparagraph\or \|\or **\or \dagger\dagger
    \or \ddagger\ddagger \else\@ctrerr\fi}}
\makeatother

\usepackage{color}
\usepackage{xcolor}
\definecolor{darkgreen}{rgb}{0.0, 0.5, 0.0}

\usepackage{colortbl}

\usepackage{lipsum}

\newcommand\nocell[2]{\multicolumn{#1}{#2}{}}
\usepackage{makecell}

\newlength{\Oldarrayrulewidth}

\usepackage{tikz}
\usetikzlibrary{fit,positioning}
\usetikzlibrary{calc}

\title{Large-Scale Unsupervised Deep Representation Learning for Brain Structure}
\titlerunning{sBMRI-Net}
\authorrunning{Anonymous}

\author{
Ayush Jaiswal\textsuperscript{1\textdagger}, Dong Guo\textsuperscript{1\textdagger}, Cauligi S. Raghavendra\textsuperscript{1$\ddagger$}, Paul Thompson\textsuperscript{2\textdagger}
}
\institute{
\textsuperscript{1}Department of Computer Science, University of Southern California\\
\textsuperscript{2}USC Imaging Genetics Center, University of Southern California\\
\textsuperscript{\textdagger}\texttt{\{ajaiswal, dongguo, pthomp\}@usc.edu}, \textsuperscript{$\ddagger$}\texttt{raghu@vsoe.usc.edu}
}

\begin{document}

\maketitle

\begin{abstract}
  Machine Learning (ML) is increasingly being used for computer aided diagnosis of brain related disorders based on structural magnetic resonance imaging (MRI) data. Most of such work employs biologically and medically meaningful hand-crafted features calculated from different regions of the brain. The construction of such highly specialized features requires a considerable amount of time, manual oversight and careful quality control to ensure the absence of errors in the computational process. Recent advances in Deep Representation Learning have shown great promise in extracting highly non-linear and information-rich features from data. In this paper, we present a novel large-scale deep unsupervised approach to learn generic feature representations of structural brain MRI scans, which requires no specialized domain knowledge or manual intervention. Our method produces low-dimensional representations of brain structure, which can be used to reconstruct brain images with very low error and exhibit performance comparable to FreeSurfer features on various classification tasks.
  
  % \blfootnote{\textsuperscript{$\star$}Data  used  in  preparing  of  this  article  were obtained in part  from  the  Alzheimer’s  Disease Neuroimaging  Initiative  (ADNI)  database (\href{http://adni.loni.usc.edu}{adni.loni.usc.edu}). As  such,    investigators within the ADNI contributed to the design and implementation of ADNI and/or provided data but  did  not  participate  in  analysis  or  writing  of  this  report.  ADNI investigators are listed at: \href{http://adni.loni.usc.edu/wp-content/uploads/how_to_apply/ADNI_Acknowledgement_List.pdf}{http://adni.loni.usc.edu/wp-content/uploads/how\_to\_apply/ADNI\_Acknowledgement\_List.pdf}} \blfootnote{\textsuperscript{$\star$}Data used in preparing this article was in part obtained from the Australian Imaging Biomarkers and Lifestyle flagship study of ageing (AIBL) funded by the Commonwealth Scientific and Industrial Research Organisation (CSIRO) and available at the ADNI database (\href{http://www.loni.usc.edu/ADNI}{www.loni.usc.edu/ADNI}). AIBL researchers contributed data but did not participate in analysis or writing of this report. AIBL researchers are listed at \href{http://www.aibl.csiro.au}{www.aibl.csiro.au}.} \blfootnote{\textsuperscript{$\star$}Data collection and sharing for this project was provided in part by the International Consortium for Brain Mapping (ICBM; Principal Investigator: John Mazziotta, MD, PhD). ICBM funding was provided by the National Institute of Biomedical Imaging and BioEngineering. ICBM data are disseminated by the Laboratory of Neuro Imaging at the University of Southern California.}
\end{abstract}

\section{Introduction}

Structural brain magnetic resonance imaging (sBMRI) helps medical practitioners make effective diagnoses of disorders by allowing the visualization of characteristics of their patients' brains and the detection of abnormalities. The curation of such data, where the diagnosis has already been made, provides opportunities to use Machine Learning (ML) methods to learn models from these examples and assist medical practitioners in making future diagnoses more efficiently.% For example, ML with structural MRI has been used for identification of Autism Spectrum Disorder (ASD)~\cite{cad:autism} and Alzheimer's Disease (AD)~\cite{cad:alz_lebedev}.%,cad:alz_hosseini}.

Developing ML models for classifying a subject's brain as normal or having a certain disorder requires an appropriate feature representation of the brain. Software such as FreeSurfer~\footnote{\href{https://surfer.nmr.mgh.harvard.edu/}{https://surfer.nmr.mgh.harvard.edu}} have been traditionally used to extract summary statistics from macro-scale brain regions that are then used to train classification models~\cite{cad:alz_lebedev,cad:autism,cad:comparative}. These features typically consist of surface area, volume and thickness of various regions of the brain. Although these specialized features are automatically computed by software, manual verification and careful quality control is required to ascertain that the process is error-free and to manually correct errors when they occur. Further, such software conform to a very strong prior over the brain, a very fixed view of anatomy, which is not correct, especially in abnormal cases where the structure of brains are deformed. Moreover, such a feature representation comprises a fixed set of targeted region properties, which fails to capture other potentially important information contained in the images.

We propose an unsupervised deep representation learning approach, based on convolutional autoencoders (CAEs)~\cite{gen:masci}, to learn low-dimensional representations of brain structure from sBMRI scans. We present three CAE models that we employed in our work: CAE-staged, CAE-joint and CAE-3D. CAE-staged and CAE-joint treat 3D brain images as a series of 2D frames along the Z-axis. CAE-staged comprises two separate CAEs: one learns frame-level representations and the other learns a combined representation of the brain structure from the frame-level encodings. We merge the two autoencoders of CAE-staged together to create the CAE-joint network for end-to-end representation learning. CAE-3D operates on the 3D brain images using 3D operations to directly learn latent embeddings. Large amounts of data are required to train these networks due to their massive number of parameters, as is true with most deep learning models. However, most publicly available sBMRI datasets are relatively very small, on the spectrum of tens to a few thousands of images. Hence, we combine data from nine different sources to create a common dataset for unsupervised representation learning. \emph{The diversity induced from pooling data from multiple sources also makes the learned features more informative and robust}. Features learned using our approach take considerably less time to construct and have performance comparable to those calculated using FreeSurfer (FS) on binary classification tasks based on identifying patients as healthy, having AD, having mild cognitive impairment (MCI), or having Autism Spectrum Disorder (ASD).

In the following sections, we present the data and preprocessing methods, conceptual understanding of CAEs, our CAE architectures, and their qualitative and quantitative analyses. To the best of our knowledge, our work is the first large-scale (spanning nine different datasets) unsupervised deep representation learning effort for structural brain images, paving the way for a plethora of possible future work, which is further motivated by our results.

\section{Data}

We train our deep CAE models on sBMRI scans, which are static 3D images of brains meant for capturing structural details, from nine different sources:

\begin{itemize}
% \footnotesize
    \item Alzheimer's Disease Neuroimaging Initiative (ADNI)~\footnote{\href{http://www.adni-info.org}{www.adni-info.org}}~\footnote{\href{http://adni.loni.usc.edu}{adni.loni.usc.edu}} - for the study of mild cognitive impairment (MCI) and early Alzheimer's disease (AD).
    \item Australian Imaging, Biomarker \& Lifestyle Flagship Study of Ageing (AIBL) - for the study of AD~\cite{data:aibl}.
    \item Open Access Series of Imaging Studies (OASIS) - for the study of AD. The data is made available by the Washington University Alzheimer's Disease Research Center, Dr. Randy Buckner at the Howard Hughes Medical Institute (HHMI) at Harvard University, the Neuroinformatics Research Group (NRG) at Washington University School of Medicine, and the Biomedical Informatics Research Network (BIRN).
    \item Autism Brain Imaging Data Exchange (ABIDE-I)~\footnote{\href{http://www.aibl.csiro.au}{www.aibl.csiro.au}} - for the study of Autism Spectrum Disorder (ASD).
    \item Brainomics/Localizer - for the study of inter-subject variability along different modalities of brain imaging~\cite{data:brainomics_1,data:brainomics_2}.
    \item Human Connectome Project (HCP) - for the study of various functions~\cite{data:hcp}.
    \item International Consortium for Brain Mapping (ICBM)~\footnote{\href{http://www.loni.usc.edu/ICBM}{www.loni.usc.edu/ICBM}} - for the development of a probabilistic reference system for the human brain.
    \item Northwestern University Schizophrenia Data and Software Tool~\footnote{\href{http://central.xnat.org/REST/projects/NUDataSharing}{http://central.xnat.org/REST/projects/NUDataSharing}} (NUSDAST) - for the study of Schizophrenia.
    \item Parkinson’s Progression Markers Initiative (PPMI)~\footnote{\href{http://www.ppmi-info.org}{www.ppmi-info.org}}~\footnote{\href{http://www.ppmi-info.org/data}{www.ppmi-info.org/data}} - for the study of the progression of Parkinson's disease.
\end{itemize}

\noindent\textbf{Data Preprocessing.} The combined dataset consists of $15,485$ images. A typical sBMRI scan is a $256 \times 256 \times 256$ image containing the subject's head. We use the skull-stripping functionality of FreeSurfer to extract only the brain from each image. We crop all the resulting images to the central $200 \times 200 \times 200$ bounding box based on our observation that this only removes empty space around the brains. We rotate each image by $-10$, $-5$, $+5$ and $+10$ degrees about each of the three axes, and translate both the original and the rotated images by $-5$ and $+5$ voxels along each of the three axes. This gives us a total of $1,409,135$ images, further enhancing the data size and making our models invariant to translation and rotation. We use $12,387$ of the original images and their augmented versions ($1,127,217$ in total) to train our models. We downsample all the images to $100 \times 100 \times 100$ dimensions to reduce the memory requirement for training.

\section{Deep Representation Learning Models}

\subsection{Convolutional Autoencoders}
\label{sec:cae}

An autoencoder~\cite{gen:hinton} (AE) is a neural network composed of an \textit{encoder}, which transforms data into its latent representation, and a \textit{decoder}, which reconstructs data from its encoding. The model is trained to minimize the reconstruction loss. Convolutional Autoencoders (CAEs)~\cite{gen:masci} effectively capture frequently occurring local features through parameter sharing across the input by employing convolution layers in the encoder and deconvolution layers in the decoder. We use max-pooling layers in CAE encoders to gradually downsample data and unpooling layers in CAE decoders for progressive upsampling. We use the rectified linear activation function (described by $y = \max (x, 0) $) at all the layers of our networks, and batch-normalization to speed up the training process.

\begin{figure}
\vspace{-10pt}
\centering
\includegraphics[trim={0 5.5cm 0 0.5cm},clip,width=0.8\textwidth]{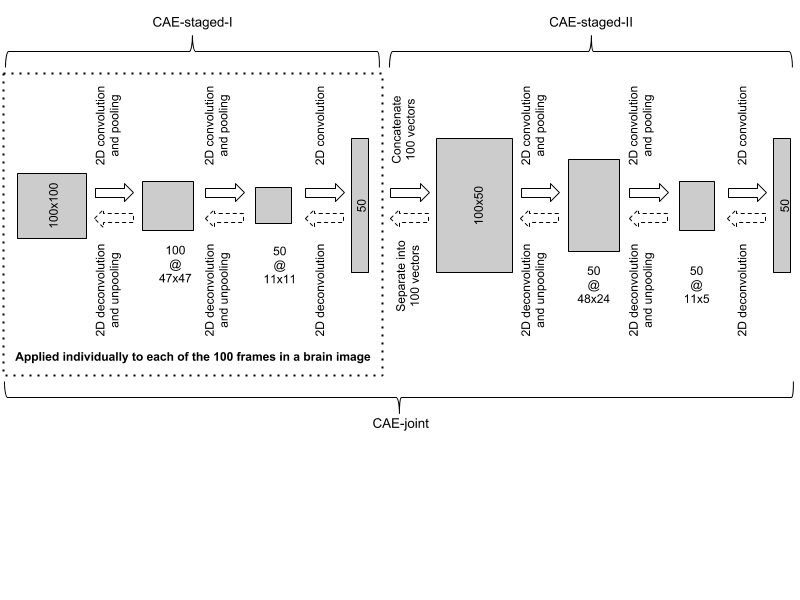}
\caption{\label{fig:arch_2d} Architecture of CAE-staged and CAE-joint models. The left half of the figure is CAE-staged-I and the right half is CAE-staged-II, as marked. Merging the layers of the two networks, we get CAE-joint as the entire graph in the figure.}
\vspace{-10pt}
\end{figure}

\begin{figure}
\centering
\includegraphics[trim={0 9.5cm 0 1.75cm},clip,width=0.8\textwidth]{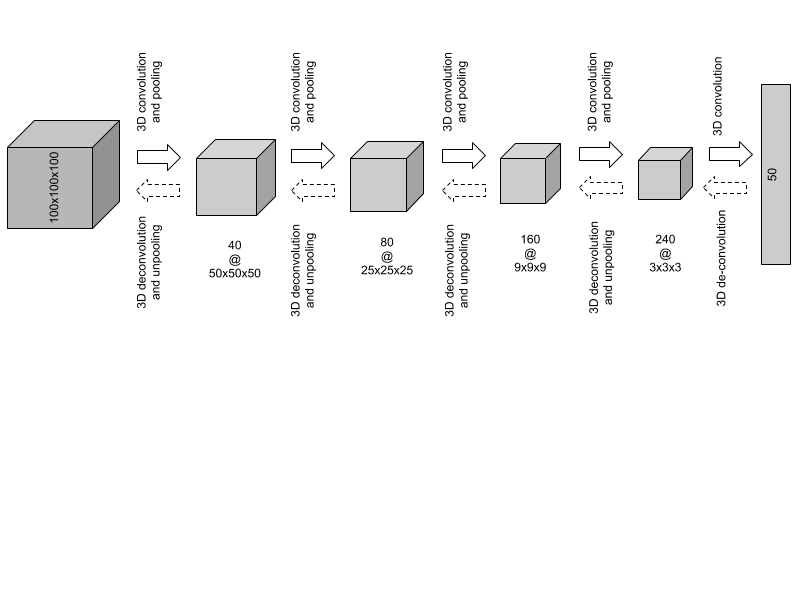}
\caption{\label{fig:arch_3d} Architecture of the CAE-3D model. All the operations are done in 3D. Hence, intermediate data for every image are 4D tensors.}
\vspace{-10pt}
\end{figure}

\subsection{CAE Architectures}

% We present three CAE models: CAE-staged and CAE-joint, which treat the 3D images as a series of 2D frames along the Z-axis, and CAE-3D, which works on each 3D image as a whole using 3D operations. We train all our models to learn $50$-dimensional representations of 3D brain images.

\noindent\textbf{CAE-staged.} This model consists of two separate CAEs: CAE-staged-I and CAE-staged-II. Figure~\ref{fig:arch_2d} describes their complete architectures as the left and the right halves of the entire graph. We first train CAE-staged-I to reconstruct any given frame so that it learns $50$-dimensional frame-level representations. We then stack the $50$-dimensional representations of all the $100$ frames to create a $(100 \times 50)$ 2D representation of each brain. We train CAE-staged-II to learn a $50$-dimensional latent representation of entire brains that can encode and reconstruct these $(100 \times 50)$ intermediate encodings.

% \vspace{5pt}
\noindent\textbf{CAE-joint.} We create the CAE-joint model by merging CAE-staged-I and CAE-staged-II into a single network, as shown in Figure~\ref{fig:arch_2d}. We reshape $50$-dimensional frame-level encodings into $(100 \times 50)$ 2D brain representations in the encoder and the reconstructed $(100 \times 50)$ representations into reconstructed $50$-dimensional frame-level encodings in the decoder. We use the learned weights from CAE-staged-I and CAE-staged-II as initial weights while training CAE-joint. CAE-joint also produces $50$-dimensional encodings for entire brains.

% \vspace{5pt}
\noindent\textbf{CAE-3D.} This model is a single CAE which treats each brain as a 3D image. It uses the 3D versions of convolution, deconvolution, pooling and unpooling layers. Figure~\ref{fig:arch_3d} describes the complete architecture of CAE-3D. We train it to directly learn a $50$-dimensional representation of brain images that can encode and reconstruct $100 \times 100 \times 100$ brain images wholly.

\begin{figure}
\vspace{-10pt}
\captionsetup[subfigure]{aboveskip=2pt,belowskip=1.5pt}
\centering
\begin{subfigure}{0.4\textwidth}
\centering
\includegraphics[width = 0.9\linewidth]{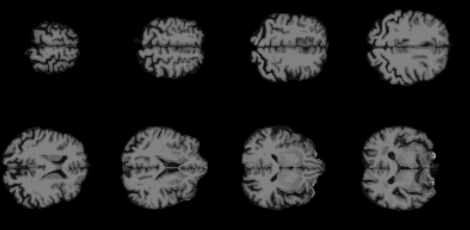}
\caption{Original brain image} \label{fig:1a}
\end{subfigure}
\begin{subfigure}{0.4\textwidth}
\centering
\includegraphics[width = 0.9\linewidth]{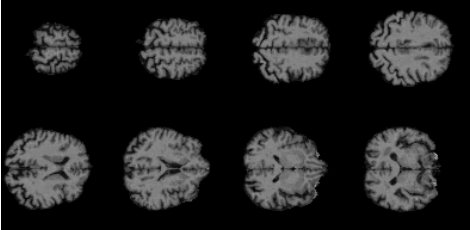}
\caption{CAE-staged-I reconstruction} \label{fig:1b}
\end{subfigure}
\begin{subfigure}{0.4\textwidth}
\centering
\includegraphics[width = 0.9\linewidth]{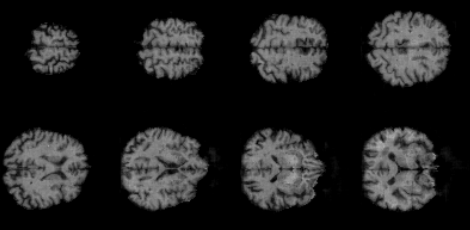}
\caption{CAE-joint reconstruction} \label{fig:1c}
\end{subfigure}
\begin{subfigure}{0.4\textwidth}
\centering
\includegraphics[width = 0.9\linewidth]{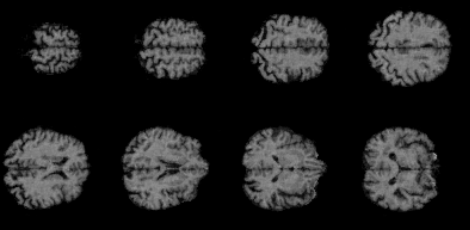}
\caption{CAE-3D reconstruction} \label{fig:1d}
\end{subfigure}
\caption{\label{recon} 2D frames in original and reconstructed brain images.}
\vspace{-15pt}
\end{figure}

\section{Experimental Evaluation}

\subsection{Qualitative Analysis}

The original brain images have voxel values between $0$ and $255$. The voxel-wise reconstruction mean \emph{squared} errors of CAE-staged-I, CAE-joint and CAE-3D on validation data are $4.43$, $10.55$ and $8.01$, respectively. We visualize reconstructed brain images by plotting a few 2D frames from the 3D tensors. Figure~\ref{recon} shows frames from original and reconstructed tensors generated by CAE-staged-I (first half of CAE-staged), CAE-joint and CAE-3D models. It can be seen that in all the three models, the learned representations can be used to generate high quality reconstructions of brain images.

To understand how our models extract features from 3D brain images, we infer and visualize the saliency maps of the nodes in the 50-dimensional brain representation layers of our models using the publicly available \texttt{keras-vis}~\footnote{\href{https://github.com/raghakot/keras-vis}{https://github.com/raghakot/keras-vis}} library. The saliency map of a node is calculated by taking the absolute value of the partial derivative of the node's value with respect to the input features. Hence, it indicates the magnitude of the effect of a voxel's perturbation on the node's activation.  Figures~\ref{fig:saliency_staged},\ref{fig:saliency_joint} and \ref{fig:saliency_3d} show the saliency maps of a few nodes in the embedding layers of CAE-staged, CAE-joint and CAE-3D, respectively. Each row in these figures corresponds to five frames of the saliency map of a single embedding node. The results show that all the three models capture features from highly-complex 3D substructures of brain images. The maps also reveal that each embedding node has different localized attention regions, which it accounts for in the information it captures. The saliency maps of the three models look dissimilar, indicating that they process brain images differently. The attention regions of CAE-3D look relatively more localized compared to those of CAE-staged and CAE-joint, indicating its stronger focus on local details.

\begingroup
\newcommand{\imscale}[0]{0.8}
\begin{figure}
\vspace{-10pt}
    \centering
    \begin{subfigure}{\textwidth}
    \centering
    \includegraphics[trim={0 1.25cm 0 0},clip,width=\imscale\linewidth]{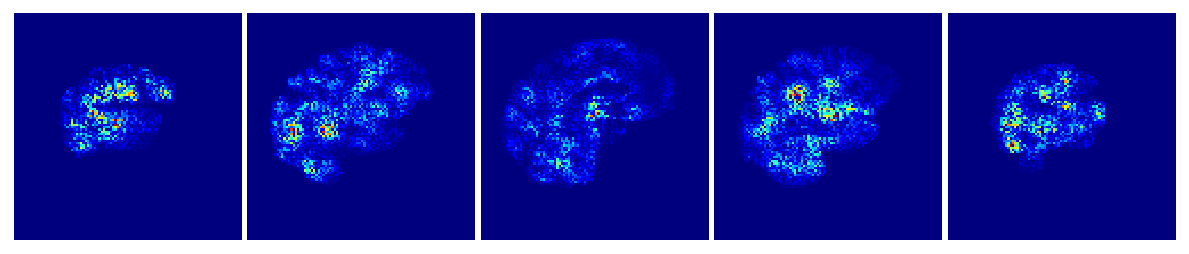}
    \end{subfigure}
    \begin{subfigure}{\textwidth}
    \centering
    \includegraphics[trim={0 1.25cm 0 0},clip,width=\imscale\linewidth]{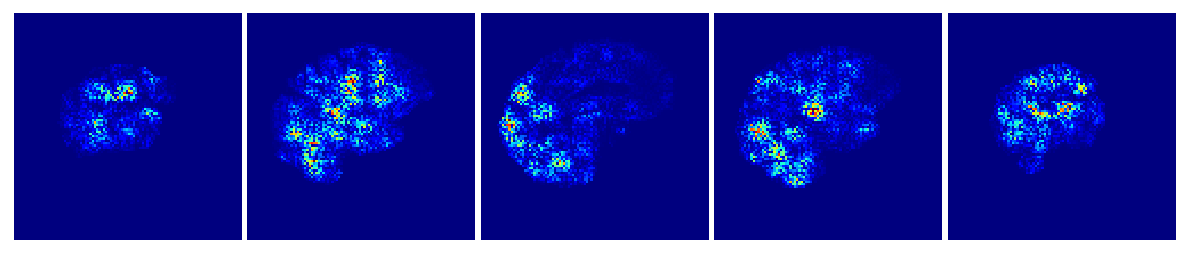}
    \end{subfigure}
    \caption{Saliency map of two nodes in the embedding layer - CAE-staged}
    \label{fig:saliency_staged}
\vspace{-25pt}
\end{figure}
\endgroup

\begingroup
\newcommand{\imscale}[0]{0.8}
\begin{figure}
    \centering
    \begin{subfigure}{\textwidth}
    \centering
    \includegraphics[trim={0 1.25cm 0 0},clip,width=\imscale\linewidth]{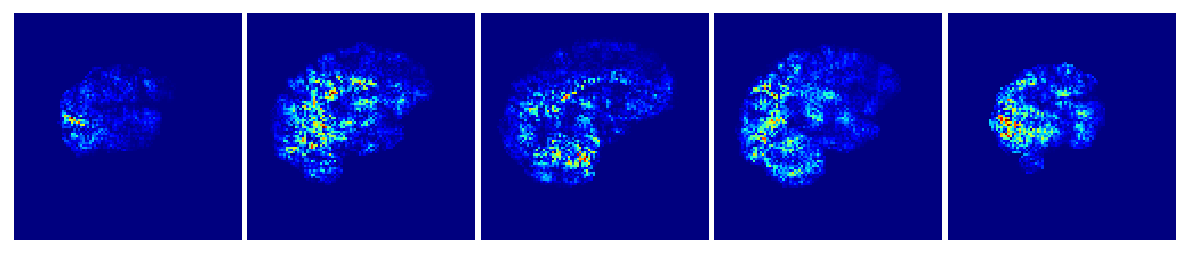}
    \end{subfigure}
    \begin{subfigure}{\textwidth}
    \centering
    \includegraphics[trim={0 1.25cm 0 0},clip,width=\imscale\linewidth]{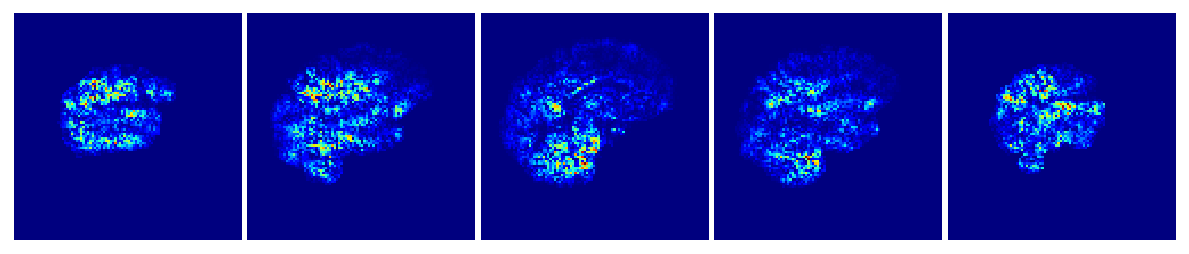}
    \end{subfigure}
    \caption{Saliency map of two nodes in the embedding layer - CAE-joint}
    \label{fig:saliency_joint}
\end{figure}
\vspace{-25pt}
\endgroup

\begingroup
\newcommand{\imscale}[0]{0.8}
\begin{figure}
    \centering
    \begin{subfigure}{\textwidth}
    \centering
    \includegraphics[trim={0 1.25cm 0 0},clip,width=\imscale\linewidth]{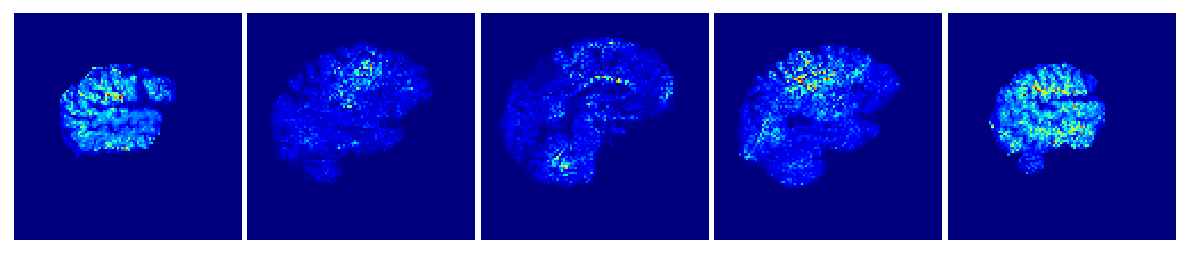}
    \end{subfigure}
    \begin{subfigure}{\textwidth}
    \centering
    \includegraphics[trim={0 1.25cm 0 0},clip,width=\imscale\linewidth]{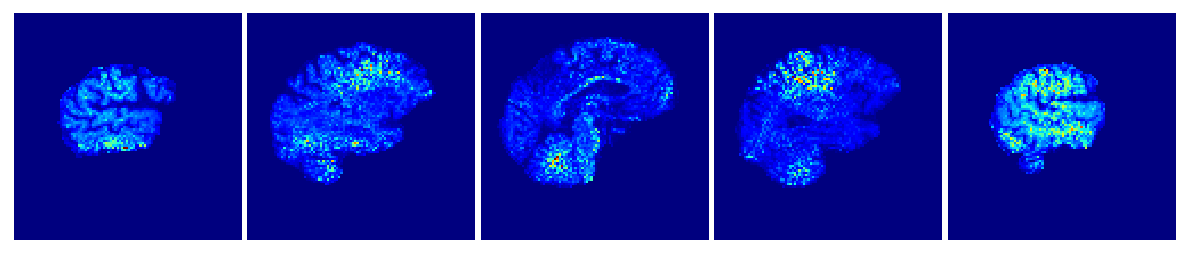}
    \end{subfigure}
    \caption{Saliency map of two nodes in the embedding layer - CAE-3D}
    \label{fig:saliency_3d}
\vspace{-15pt}
\end{figure}
\endgroup

% \begin{figure}
% \centering
% \captionsetup[subfigure]{aboveskip=2pt,belowskip=1.5pt}
% \begin{subfigure}{0.3\textwidth}
% \centering
% \includegraphics[width = 0.8\linewidth]{img/caestage_small.png}
% \caption{CAE-staged-II filter} \label{fig:2b}
% \end{subfigure}
% \begin{subfigure}{0.3\textwidth}
% \centering
% %\includegraphics[width = \linewidth]{img/caejoint_small.png}
% %\includegraphics[width = \linewidth]{img/cae_joint_selected.png}
% \includegraphics[width = 0.8\linewidth]{img/cae_joint_selected_contrast.png}
% \caption{CAE-joint filter} \label{fig:2c}
% \end{subfigure}
% \begin{subfigure}{0.3\textwidth}
% \centering
% %\includegraphics[width = \linewidth]{img/cae3d_small.png}
% %\includegraphics[width = \linewidth]{img/cae3d_selected.png}
% \includegraphics[width = 0.8\linewidth]{img/cae3d_selected_contrast.png}
% \caption{CAE-3D filter} \label{fig:2d}
% \end{subfigure}
% \caption{\label{filters} Frames from input tensors that maximize activation of representation layer hidden nodes in CAE-staged-II, CAE-joint, and CAE-3D models. Each subplot of 9 frames corresponds to one node in the representation layer.}
% \vspace{-15pt}
% \end{figure}

\subsection{Quantitative Analysis}

We compare the learned CAE embeddings with the features extracted using FS on four classification tasks. Thus, we evaluate four feature representations: (1) FreeSurfer (FS), (2) CAE-staged (CAES), (3) CAE-joint (CAEJ) and (4) CAE-3D. The classification tasks are derived from two datasets: ADNI and ABIDE-I. The ADNI dataset has $426$ healthy subjects (H-ADNI), $348$ subjects with AD and $255$ subjects with MCI. The ABIDE-I dataset has $421$ healthy subjects (H-ABIDE) and $430$ subjects with ASD. The classification tasks we evaluate on are: (1) H-ADNI vs. AD, (2) H-ADNI vs. MCI, (3) AD vs. MCI, and (4) H-ABIDE vs. ASD. We train a Logistic Regression classifier and a Random Forest classifier using each feature representation for each classification task. We use Area Under the Receiver Operating Characteristic curve (AUROC)~\cite{gen:auroc} as the evaluation metric. We use $80\%$ of the data for training and $20\%$ for testing, split randomly. We summarize the results of our experiments on test data in Table~\ref{table:quant}. We see that the embeddings learned by our models show performance comparable to traditionally used FS features on all the aforementioned classification tasks, with CAE-3D performing slightly better than CAE-joint.

\vspace{5pt}
\noindent\textbf{Timing Analysis.} Feature construction using FreeSurfer takes $20-47$ hours for each  image~\footnote{\href{https://surfer.nmr.mgh.harvard.edu/fswiki/ReconAllRunTimes}{https://surfer.nmr.mgh.harvard.edu/fswiki/ReconAllRunTimes}}. In contrast, CAE-staged and CAE-joint take $0.55s$, and CAE-3D takes $0.45s$ on average to generate the latent embedding of each brain image.

\begin{table}
\vspace{-25pt}
\setlength\tabcolsep{2pt}
\setlength\extrarowheight{1pt}
\centering
\small
\caption{AUROC with Logistic Regression and Random Forest. Colors indicate similarity of scores obtained using the same classifier for each classification task. Darker colors show higher scores.}
 \begin{tabular}{ l@{\hskip 10pt} c  c  c  c@{\hskip 10pt} c  c  c  c } 
%  \nocell{9}{c}\\ %\cmidrule[1pt](r{8pt}){2-5} \cmidrule[1pt]{6-9}
 \nocell{1}{c} & \multicolumn{4}{c}{\textbf{Logistic Regression}} & \multicolumn{4}{c}{\textbf{Random Forest}}\\  \cmidrule[1pt](l{-2pt}r{8pt}){2-5} \cmidrule[1pt](l{-2pt}){6-9}
 \textbf{Classification} & \textbf{FS} & \textbf{CAES} & \textbf{CAEJ} & \textbf{CAE-3D} & \textbf{FS} & \textbf{CAES} & \textbf{CAEJ} & \textbf{CAE-3D}\\
 \cmidrule[1pt](r{8pt}){1-1} \cmidrule[1pt](l{-2pt}r{8pt}){2-5} \cmidrule[1pt](l{-2pt}){6-9}
 H-ADNI / AD & \cellcolor{blue!35}0.81 & \cellcolor{blue!10}0.67 & \cellcolor{blue!35}0.82 & \cellcolor{blue!35}0.81 & \cellcolor{darkgreen!35}0.86 & \cellcolor{darkgreen!10}0.64 & \cellcolor{darkgreen!25}0.80 & \cellcolor{darkgreen!33}0.83 \\
 AD / MCI & \cellcolor{blue!30}0.71 & \cellcolor{blue!25}0.67 & \cellcolor{blue!35}0.76 & \cellcolor{blue!30}0.72 & \cellcolor{darkgreen!35}0.77 & \cellcolor{darkgreen!25}0.70 & \cellcolor{darkgreen!33}0.73 & \cellcolor{darkgreen!33}0.73 \\
 H-ADNI / MCI & \cellcolor{blue!35}0.77 & \cellcolor{blue!33}0.75 & \cellcolor{blue!35}0.76 & \cellcolor{blue!35}0.76 & \cellcolor{darkgreen!35}0.81 & \cellcolor{darkgreen!25}0.71 & \cellcolor{darkgreen!33}0.77 & \cellcolor{darkgreen!35}0.80 \\
 H-ABIDE / ASD & \cellcolor{blue!35}0.60  & \cellcolor{blue!30}0.57 & \cellcolor{blue!30}0.57 & \cellcolor{blue!35}0.60 & \cellcolor{darkgreen!35}0.65 & \cellcolor{darkgreen!35}0.64 & \cellcolor{darkgreen!33}0.63 & \cellcolor{darkgreen!35}0.66  \\
 \cmidrule[1pt](r{8pt}){1-1} \cmidrule[1pt](l{-2pt}r{8pt}){2-5} \cmidrule[1pt](l{-2pt}){6-9}
 \end{tabular}
 \label{table:quant}
 \vspace{-35pt}
\end{table}

\section{Related Work}

% Hinton and Salakhutdinov~\cite{gen:hinton} introduced autoencoders as a method for non-linear dimensionality reduction. They showed how neural networks can be trained to reconstruct their input to learn low-dimensional latent representations. Zeiler et al.~\cite{gen:zeiler} introduced the concept of deconvolutional layers in a fully deconvolutional network to learn hidden representations of data. Masci et al.~\cite{gen:masci} introduced Convolutional Autoencoders (CAEs) as a variant of autoencoders with convolutional layers to take advantage of repeating spatial structure occurring in images. The CAE framework has been used extensively to learn latent representation of images~\cite{gen:cae_rep_1,gen:cae_rep_2}.

G\"{u}\c{c}l\"{u} and Gerven~\cite{drl:umut} showed that unsupervised feature learning from functional brain MRI data improves human brain activity prediction in response to natural images. Brosch et al.~\cite{brosch2013manifold} proposed a method for manifold learning of brain images in the ADNI dataset using Deep Belief Networks (DBNs) composed of convolutional Restricted Boltzmann Machines. However, their method is not end-to-end trainable and, hence, not scaleable. Plis et al.~\cite{drl:plis} presented the use of DBNs in a Constraint Satisfaction Problem framework to learn latent embeddings of gray matter images extracted from sBMRI scans on two datasets \emph{separately}. Our work is different from theirs as we propose a \emph{large-scale} deep unsupervised approach to learn latent representations of brain structure from sBMRI scans, spanning nine different datasets, \emph{without the need for first extracting gray matter images}.

% Greenstein et al~\cite{cad:schiz_greenstein} present a method to classify patients with early onset of schizophrenia. Dinov et al.~\cite{cad:parkinsons} use similar features to detect Parkinson's disease. 
%Hosseini-Asl et al.~\cite{cad:alz_hosseini} use a 3D Convolutional Neural Network for the detection of AD. Retico et al.~\cite{cad:autism} provide a review of the use of ML in detecting ASD in patients using features derived from their sBMRI scans.

\section{Conclusion}

We presented an approach for large-scale deep unsupervised representation learning for sBMRI data incorporating models based on CAEs: CAE-staged, CAE-joint and CAE-3D. Features learned using our approach can reconstruct brain images with very low error and show performance comparable to FS features on classification tasks. Feature encoding using our method takes \emph{considerably less time} compared to FS features, \emph{while employing no specialized domain knowledge}. Our models \emph{do not have a fixed view of brain anatomy} and can be made \emph{increasingly general with diversified training data}. The proposed models are adaptive, i.e., the learned features can be improved by training with more data.

% % Uncomment below for camera-ready
% \vspace{5pt}
{
\begingroup
% \scriptsize
\noindent\textbf{Acknowledgments.} The collection and sharing of data used in this  work  was funded  by  NIH Grant  U01 AG024904, Department  of  Defense  award  number  W81XWH-12-2-0012, P50 AG05681, P01 AG03991, R01 AG021910, P20 MH071616, U24 RR021382, NIMH K23MH087770, the McDonnell Center for Systems Neuroscience at Washington University, 16 NIH Institutes and Centers that support the NIH Blueprint for Neuroscience Research, National Institute of Biomedical Imaging and BioEngineering, NIMH 1R01 MH084803, the Michael J. Fox  Foundation for Parkinson’s Research and funding partners, the Leon Levy Foundation and primary support for the work by Michael P. Milham (MPM) and the INDI team was provided by gifts from Joseph P. Healy and the Stavros Niarchos Foundation to the Child Mind Institute, as well as by an NIMH award to MPM (R03MH096321). The authors also thank Daniel Moyer, Neda Jahanshad and Marc Harrison of USC Image Genetics Center for assistance with retrieving part of the data used in this project. Computation for the work described in this paper was supported by \href{http://hpcc.usc.edu}{USC's Center for High-Performance Computing}.
\endgroup
}

\begin{spacing}{0.8}
\bibliographystyle{splncs03}
\bibliography{mybibliography}
\end{spacing}

\end{document}